\documentclass[conference]{IEEEtran}

\usepackage{cite}
\ifCLASSINFOpdf
  
   \usepackage[pdftex]{graphicx}
  
\else
\fi
\usepackage{url}
\begin{document}
%
% paper title
% Titles are generally capitalized except for words such as a, an, and, as,
% at, but, by, for, in, nor, of, on, or, the, to and up, which are usually
% not capitalized unless they are the first or last word of the title.
% Linebreaks \\ can be used within to get better formatting as desired.
% Do not put math or special symbols in the title.
\title{What Drives the International Development Agenda? An NLP Analysis of the United Nations General Debate 1970-2016}

% author names and affiliations
% use a multiple column layout for up to three different
% affiliations
\author{\IEEEauthorblockN{Alexander Baturo}
\IEEEauthorblockA{School of Law and Government\\
Dublin City University\\
Email: alex.baturo@dcu.ie}
\and
\IEEEauthorblockN{Niheer Dasandi}
\IEEEauthorblockA{School of Government and Society\\
University of Birmingham\\
Email: n.dasandi@bham.ac.uk}
\and
\IEEEauthorblockN{Slava J. Mikhaylov}
\IEEEauthorblockA{Institute for Analytics and Data Science\\
Department of Government\\
University of Essex\\
Email: s.mikhaylov@essex.ac.uk}}

% conference papers do not typically use \thanks and this command
% is locked out in conference mode. If really needed, such as for
% the acknowledgment of grants, issue a \IEEEoverridecommandlockouts
% after \documentclass

% for over three affiliations, or if they all won't fit within the width
% of the page, use this alternative format:
% 
%\author{\IEEEauthorblockN{Michael Shell\IEEEauthorrefmark{1},
%Homer Simpson\IEEEauthorrefmark{2},
%James Kirk\IEEEauthorrefmark{3}, 
%Montgomery Scott\IEEEauthorrefmark{3} and
%Eldon Tyrell\IEEEauthorrefmark{4}}
%\IEEEauthorblockA{\IEEEauthorrefmark{1}School of Electrical and Computer Engineering\\
%Georgia Institute of Technology,
%Atlanta, Georgia 30332--0250\\ Email: see http://www.michaelshell.org/contact.html}
%\IEEEauthorblockA{\IEEEauthorrefmark{2}Twentieth Century Fox, Springfield, USA\\
%Email: homer@thesimpsons.com}
%\IEEEauthorblockA{\IEEEauthorrefmark{3}Starfleet Academy, San Francisco, California 96678-2391\\
%Telephone: (800) 555--1212, Fax: (888) 555--1212}
%\IEEEauthorblockA{\IEEEauthorrefmark{4}Tyrell Inc., 123 Replicant Street, Los Angeles, California 90210--4321}}

% use for special paper notices
%\IEEEspecialpapernotice{(Invited Paper)}

% make the title area
\maketitle

% As a general rule, do not put math, special symbols or citations
% in the abstract
\begin{abstract}

There is surprisingly little known about agenda setting for international development in the United Nations (UN) despite it having a significant influence on the process and outcomes of development efforts. This paper addresses this shortcoming using a novel approach that applies natural language processing techniques to countries' annual statements in the UN General Debate. Every year UN member states deliver statements during the General Debate on their governments' perspective on major issues in world politics. These speeches provide invaluable information on state preferences on a wide range of issues, including international development, but have largely been overlooked in the study of global politics. This paper identifies the main international development topics that states raise in these speeches between 1970 and 2016, and examine the country-specific drivers of international development rhetoric.
\end{abstract}

% no keywords

% For peer review papers, you can put extra information on the cover
% page as needed:
% \ifCLASSOPTIONpeerreview
% \begin{center} \bfseries EDICS Category: 3-BBND \end{center}
% \fi
%
% For peerreview papers, this IEEEtran command inserts a page break and
% creates the second title. It will be ignored for other modes.
\IEEEpeerreviewmaketitle

\section{Introduction}
Decisions made in international organisations are fundamental to international development efforts and initiatives. It is in these global governance arenas that the rules of the global economic system, which have a huge impact on development outcomes are agreed on; decisions are made about large-scale funding for development issues, such as health and infrastructure; and key development goals and targets are agreed on, as can be seen with the Millennium Development Goals (MDGs). More generally, international organisations have a profound influence on the ideas that shape international development efforts \cite{hudson2014global}.

Yet surprisingly little is known about the agenda-setting process for international development in global governance institutions. This is perhaps best demonstrated by the lack of information on how the different goals and targets of the MDGs were decided, which led to much criticism and concern about the global governance of development \cite{saith2006universal}. More generally, we know little about the types of development issues that different countries prioritise, or whether country-specific factors such as wealth or democracy make countries more likely to push for specific development issues to be put on the global political agenda. 

The lack of knowledge about the agenda setting process in the global governance of development is in large part due to the absence of obvious data sources on states' preferences about international development issues. To address this gap we employ a novel approach based on the application of natural language processing (NLP) to countries' speeches in the UN. Every September, the heads of state and other high-level country representatives gather in New York at the start of a new session of the United Nations General Assembly (UNGA) and address the Assembly in the General Debate. The General Debate (GD) provides the governments of the almost two hundred UN member states with an opportunity to present their views on key issues in international politics -- including international development. As such, the statements made during GD are an invaluable and, largely untapped, source of information on governments' policy preferences on international development over time. 

An important feature of these annual country statements is that they are not institutionally connected to decision-making in the UN. This means that governments face few external constraints when delivering these speeches, enabling them to raise the issues that they consider the most important. Therefore, the General Debate acts ``as a barometer of international opinion on important issues, even those not on the agenda for that particular session'' \cite{smith2006}. In fact, the GD is usually the first item for each new session of the UNGA, and as such it provides a forum for governments to identify like-minded members, and to put on the record the issues they feel the UNGA should address. Therefore, the GD can be viewed as a key forum for governments to put different policy issues on international agenda. 

We use a new dataset of GD statements from 1970 to 2016, \emph{the UN General Debate Corpus} (UNGDC), to examine the international development agenda in the UN \cite{baturo2017understanding}.\footnote{UNGDC is publicly available at the Harvard Dataverse at \url{http://dx.doi.org/10.7910/DVN/0TJX8Y}} Our application of NLP to these statements focuses in particular on structural topic models (STMs)\cite{roberts2013structural}. The paper makes two contributions using this approach: (1) It sheds light on the main international development issues that governments prioritise in the UN; and (2) It identifies the key country-specific factors associated with governments discussing development issues in their GD statements.

\section*{The UN General Debate and international development}

In the analysis we consider the nature of international development issues raised in the UN General Debates, and the effect of structural covariates on the level of developmental rhetoric in the GD statements. To do this, we first implement a structural topic model \cite{roberts2013structural}. This enables us to identify the key international development topics discussed in the GD. We model topic prevalence in the context of the structural covariates. In addition, we control for region fixed effects and time trend. The aim is to allow the observed metadata to affect the frequency with which a topic is discussed in General Debate speeches. This allows us to test the degree of association between covariates (and region/time effects) and the average proportion of a document discussing a topic. 

\subsection{Estimation of topic models}

We assess the optimal number of topics that need to be specified for the STM analysis. We follow the recommendations of the original STM paper and focus on exclusivity and semantic coherence measures. \cite{mimno2011optimizing} propose semantic coherence measure, which is closely related to point-wise mutual information measure posited by \cite{newman2010automatic} to evaluate topic quality. \cite{mimno2011optimizing} show that semantic coherence corresponds to expert judgments and more general human judgments in Amazon's Mechanical Turk experiments.

%%%%%%%%%%%%%%%%%%%%%%%%%%%%%%%%%%%%%%%%
\begin{figure}
\centering

\includegraphics[width=.45\textwidth]{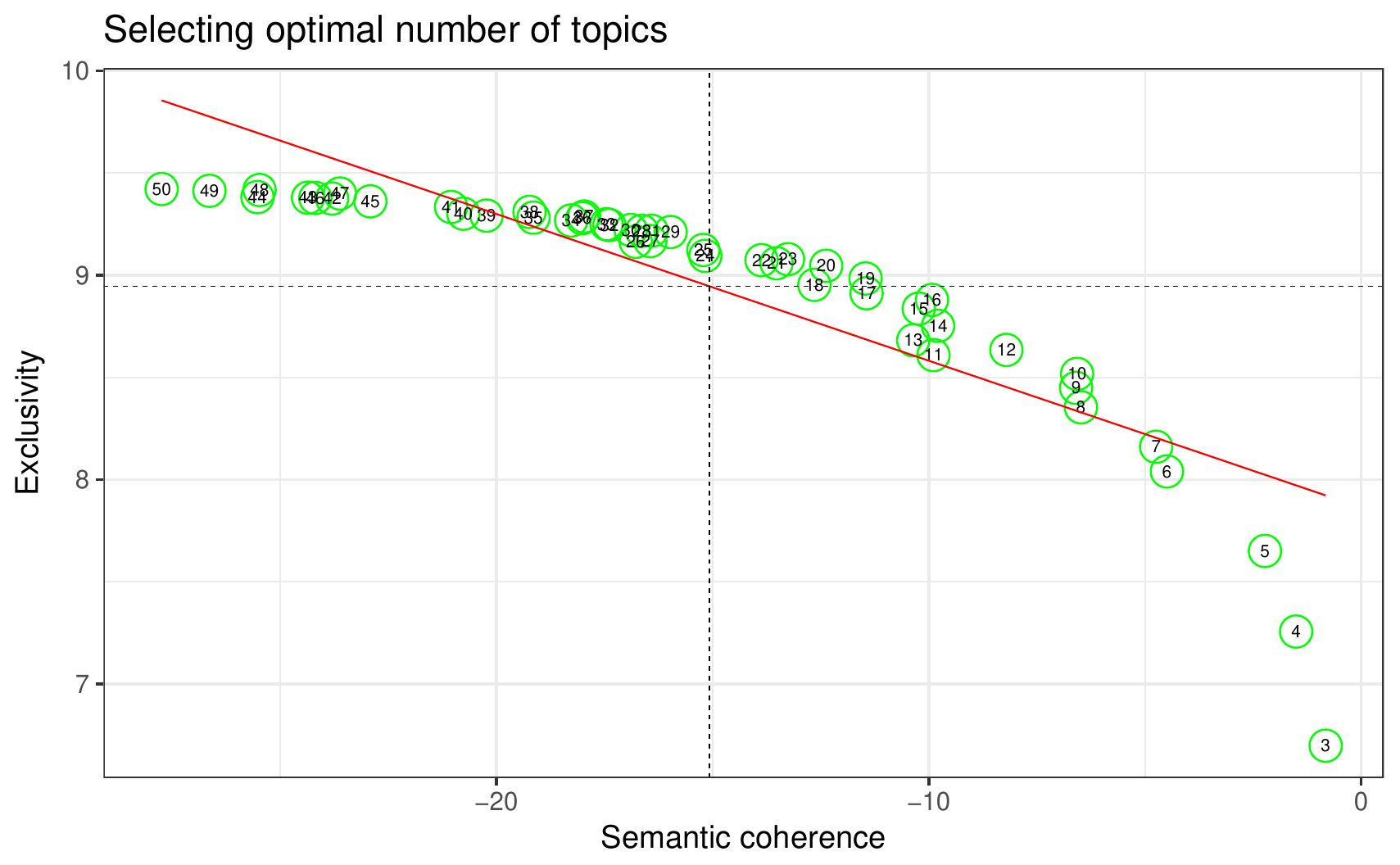} 
\caption{\emph{Optimal model search}. Semantic coherence and exclusivity results for a model search from 3 to 50 topics. Models above the regression line provide a better trade off. Largest positive residual is a 16-topic model. 
\label{fig:search}}
\end{figure}
%%%%%%%%%%%%%%%%%%%%%%%%%%%%%%%%%%%%%%%%

\begin{figure}
\centering

\includegraphics[width=.45\textwidth]{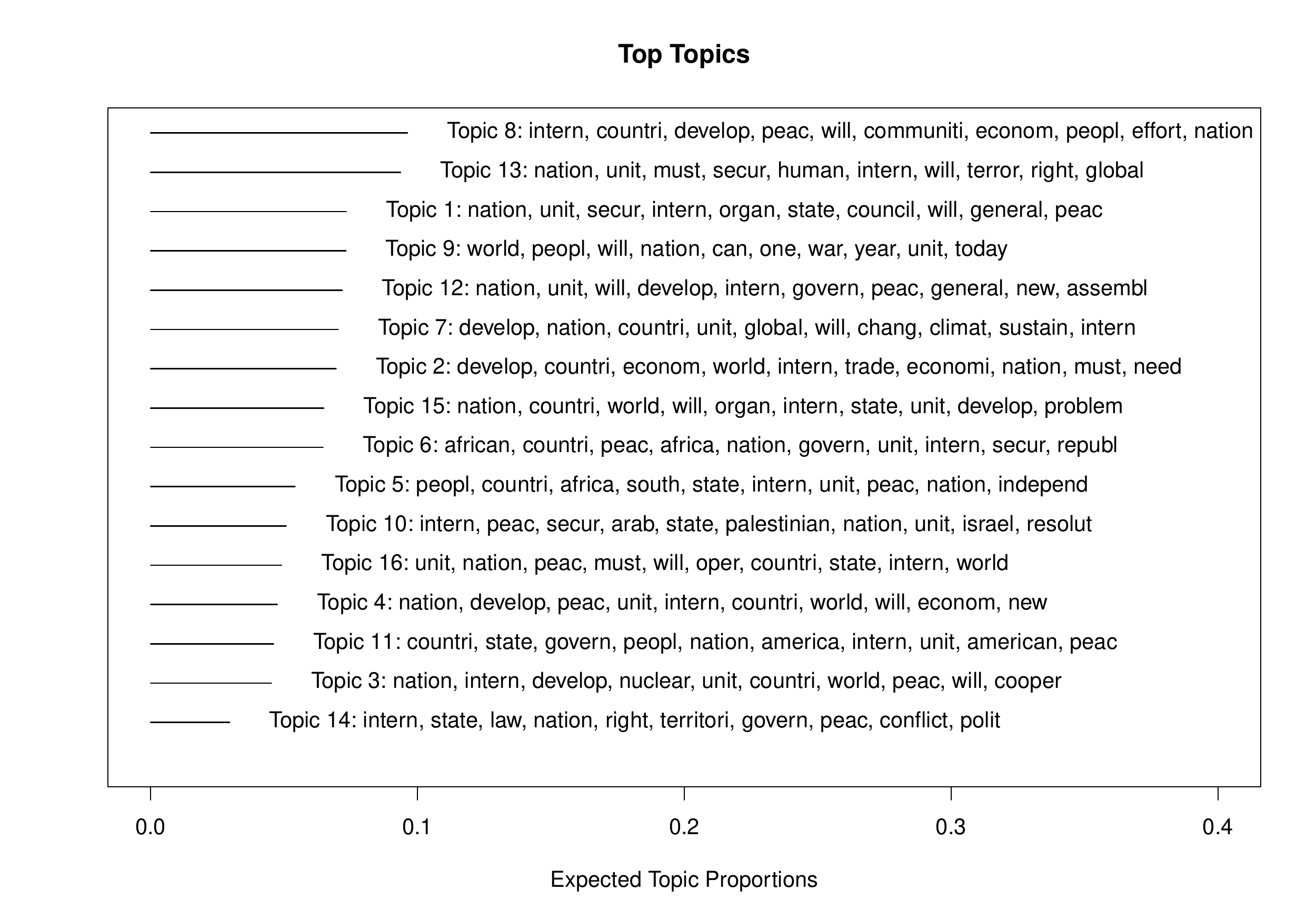} 
\caption{\emph{Topic quality}. 20 highest probability words for the 16-topic model. 
\label{fig:words}}
\end{figure}

Exclusivity scores for each topic follows \cite{bischof2012summarizing}. Highly frequent words in a given topic that do not appear very often in other topics are viewed as making that topic exclusive. Cohesive and exclusive topics are more semantically useful. Following \cite{roberts2016stm} we generate a set of candidate models ranging between 3 and 50 topics. We then plot the exclusivity and semantic coherence (numbers closer to 0 indicate higher coherence), with a linear regression overlaid (Figure \ref{fig:search}). Models above the regression line have a ``better'' exclusivity-semantic coherence trade off. We select the 16-topic model, which has the largest positive residual in the regression fit, and provides higher exclusivity at the same level of semantic coherence. 
The topic quality is usually evaluated by highest probability words, which is presented in Figure \ref{fig:words}.

%%%%%%%%%%%%%%%%%%%%%%%%%%%%%%%%%%%%%%%%
%%%%%%%%%%%%%%%%%%%%%%%%%%%%%%%%%%%%%%%%

\subsection{Topics in the UN General Debate}

Figure \ref{fig:words} provides a list of the main topics (and the highest probability words associated these topics) that emerge from the STM of UN General Debate statements. In addition to the highest probability words, we use several other measures of key words (not presented here) to interpret the dimensions. This includes the FREX metric (which combines exclusivity and word frequency), the \emph{lift} (which gives weight to words that appear less frequently in other topics), and the \emph{score} (which divides the log frequency of the word in the topic by the log frequency of the word in other topics). We provide a brief description of each of the 16 topics here.

\noindent \textbf{Topic 1} - \emph{Security and cooperation in Europe}.

The first topic is related to issues of security and cooperation, with a focus on Central and Eastern Europe. 

\noindent \textbf{Topic 2} - \emph{Economic development and the global system}.

This topic is related to economic development, particularly around the global economic system. The focus on `trade', `growth', `econom-', `product', `growth',  `financ-', and etc. suggests that Topic 2 represent a more traditional view of international development in that the emphasis is specifically on economic processes and relations.

\noindent \textbf{Topic 3} - \emph{Nuclear disarmament}.

This topic picks up the issue of nuclear weapons, which has been a major issue in the UN since its founding.

\noindent \textbf{Topic 4} - \emph{Post-conflict development}.

This topic relates to post-conflict development. The countries that feature in the key words (e.g. Rwanda, Liberia, Bosnia) have experienced devastating civil wars, and the emphasis on words such as `develop', `peace', `hope', and `democrac-' suggest that this topic relates to how these countries recover and move forward.

\noindent \textbf{Topic 5} - \emph{African independence / decolonisation}.

This topic picks up the issue of African decolonisation and independence. It includes the issue of apartheid in South Africa, as well as racism and imperialism more broadly. 

\noindent \textbf{Topic 6} - \emph{Africa}.

While the previous topic focused explicitly on issues of African independence and decolonisation, this topic more generally picks up issues linked to Africa, including peace, governance, security, and development. 

\noindent \textbf{Topic 7} - \emph{Sustainable development}.

This topic centres on sustainable development, picking up various issues linked to development and climate change. In contrast to Topic 2, this topic includes some of the newer issues that have emerged in the international development agenda, such as sustainability, gender, education, work and the MDGs.

\noindent \textbf{Topic 8} - \emph{Functional topic}.

This topic appears to be comprised of functional or process-oriented words e.g. `problem', `solution', `effort', `general', etc. 

\noindent \textbf{Topic 9} - \emph{War}.

This topic directly relates to issues of war. The key words appear to be linked to discussions around ongoing wars.

\noindent \textbf{Topic 10} - \emph{Conflict in the Middle East}.

This topic clearly picks up issues related to the Middle East -- particularly around peace and conflict in the Middle East. 

\noindent \textbf{Topic 11} - \emph{Latin America}.

This is another topic with a regional focus, picking up on issues related to Latin America. 

\noindent \textbf{Topic 12} - \emph{Commonwealth}.

This is another of the less obvious topics to emerge from the STM in that the key words cover a wide range of issues. However, the places listed (e.g. Australia, Sri Lanka, Papua New Guinea) suggest the topic is related to the Commonwealth (or former British colonies). 

\noindent \textbf{Topic 13} - \emph{International security}.

This topic broadly captures international security issues (e.g. terrorism, conflict, peace) and in particularly the international response to security threats, such as the deployment of peacekeepers. 

\noindent \textbf{Topic 14} - \emph{International law}.

This topic picks up issues related to international law, particularly connected to territorial disputes.

\noindent \textbf{Topic 15} - \emph{Decolonisation}.

This topic relates more broadly to decolonisation. As well as specific mention of decolonisation, the key words include a range of issues and places linked to the decolonisation process. 

\noindent \textbf{Topic 16} - \emph{Cold War}.

This is another of the less tightly defined topics. The topics appears to pick up issues that are broadly related to the Cold War. There is specific mention of the Soviet Union, and detente, as well as issues such as nuclear weapons, and the Helsinki Accords.

\begin{figure}
\centering
\includegraphics[width=.45\textwidth]{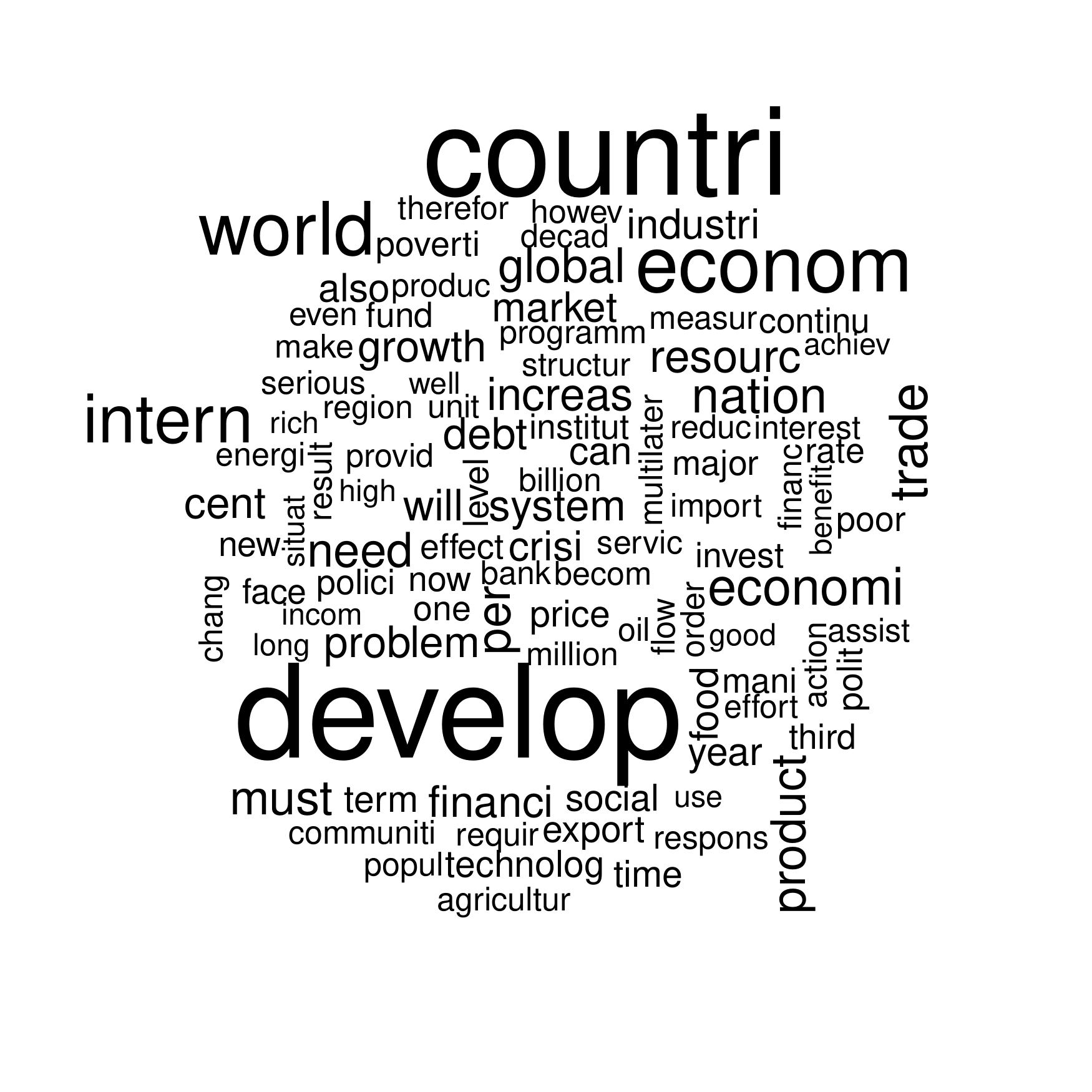}\\
\includegraphics[width=.45\textwidth]{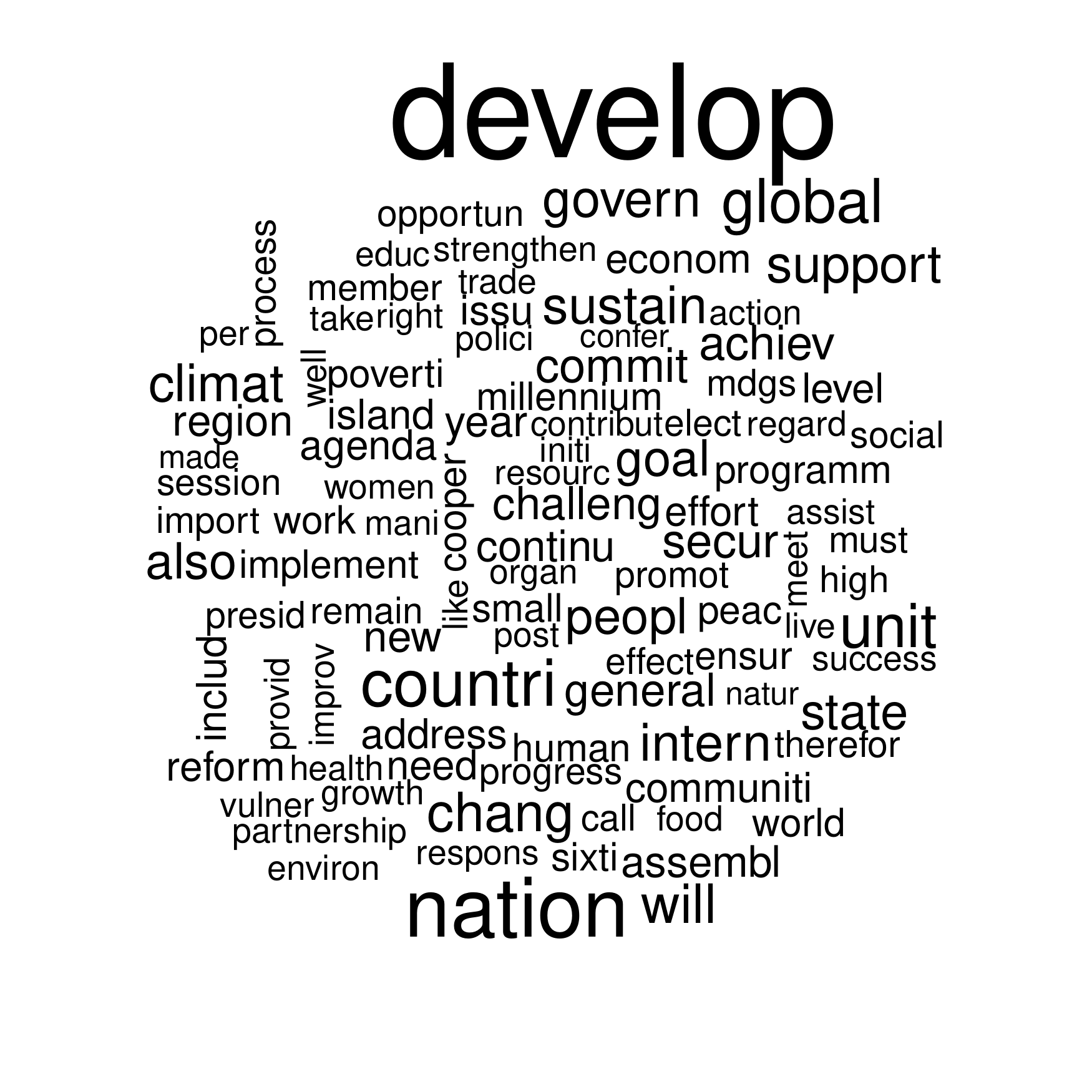} 
\caption{\emph{Topic content}. 50 highest probability words for the 2nd and 7th topics. 
\label{fig:wordclouds}}
\end{figure}

Based on these topics, we examine Topic 2 and Topic 7 as the principal ``international development'' topics. While a number of other topics -- for example post-conflict development, Africa, Latin America, etc. -- are related to development issues, Topic 2 and Topic 7 most directly capture aspects of international development. We consider these two topics more closely by contrasting the main words linked to these two topics. In Figure \ref{fig:wordclouds}, the word clouds show the 50 words most likely to mentioned in relation to each of the topics.

%%%%%%%%%%%%%%%%%%%%%%%%%%%%%%%%%%%%%%%%
%%%%%%%%%%%%%%%%%%%%%%%%%%%%%%%%%%%%%%%%

The word clouds provide further support for Topic 2 representing a more traditional view of international development focusing on economic processes. In addition to a strong emphasis on 'econom-', other key words, such as `trade', `debt', `market', `growth', `industri-', `financi-', `technolog-', `product', and `argicultur-', demonstrate the narrower economic focus on international development captured by Topic 2. In contrast, Topic 7 provides a much broader focus on development, with key words including `climat-', `sustain', `environ-', `educ-', `health', `women', `work', `mdgs', `peac-', `govern-', and `right'. Therefore, Topic 7 captures many of the issues that feature in the recent Sustainable Development Goals (SDGs) agenda \cite{waage2015governing}. 

\begin{figure}
\centering

\includegraphics[width=.45\textwidth]{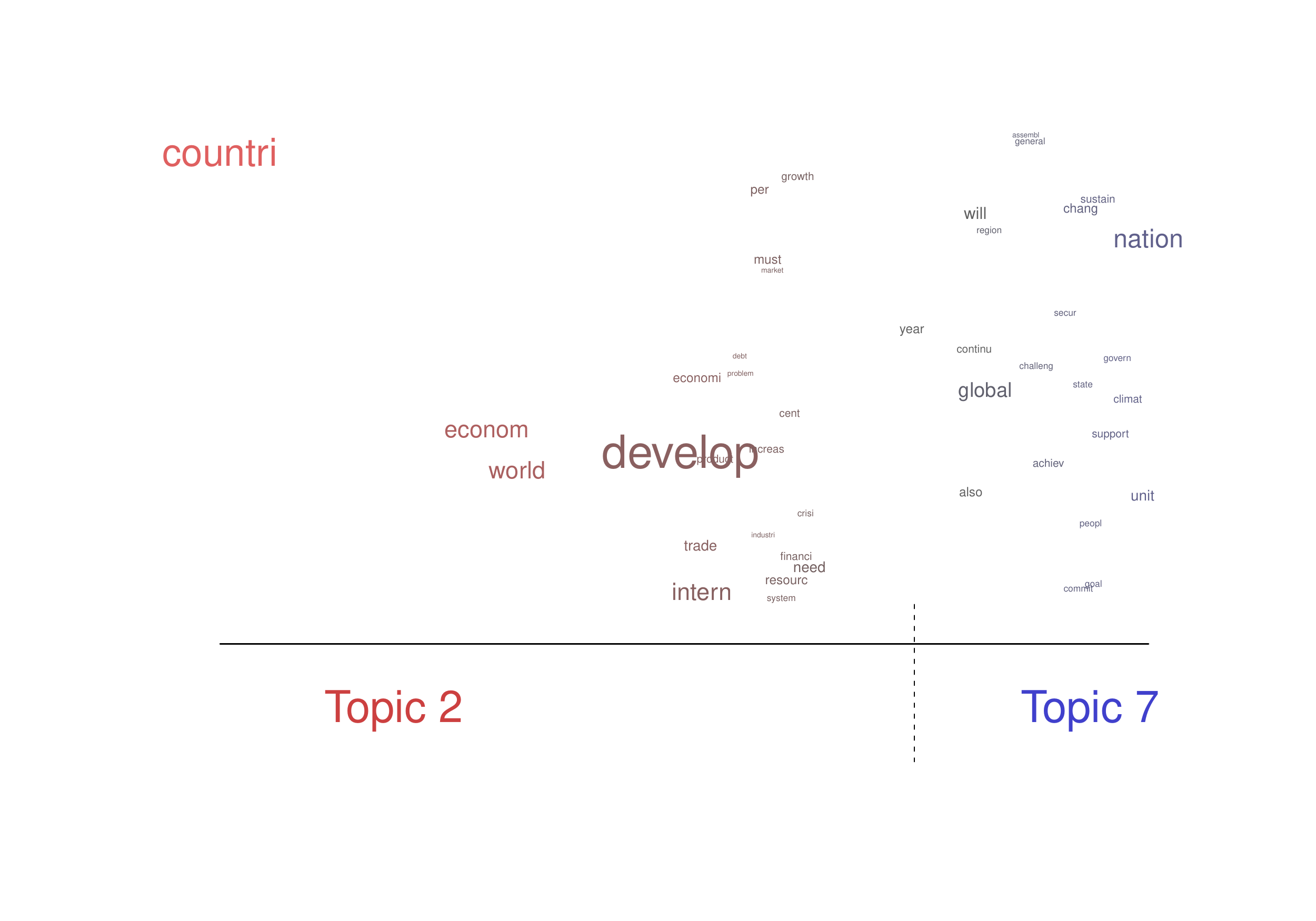} 
\caption{\emph{Comparing Topics 2 and 7 quality}. 50 highest probability words contrasted between Topics 2 and 7. 
\label{fig:perspectives}}
\end{figure}

Figure \ref{fig:perspectives} calculates the difference in probability of a word for the two topics, normalized by the maximum difference in probability of any word between the two topics. The figure demonstrates that while there is a much high probability of words, such as `econom-', `trade', and even `develop-' being used to discuss Topic 2; words such as `climat-', `govern-', `sustain', `goal', and `support' being used in association with Topic 7. This provides further support for the Topic 2 representing a more economistic view of international development, while Topic 7 relating to a broader sustainable development agenda.

%%%%%%%%%%%%%%%%%%%%%%%%%%%%%%%%%%%%%%%%
%%%%%%%%%%%%%%%%%%%%%%%%%%%%%%%%%%%%%%%%

\begin{figure}
\centering

\includegraphics[width=.45\textwidth]{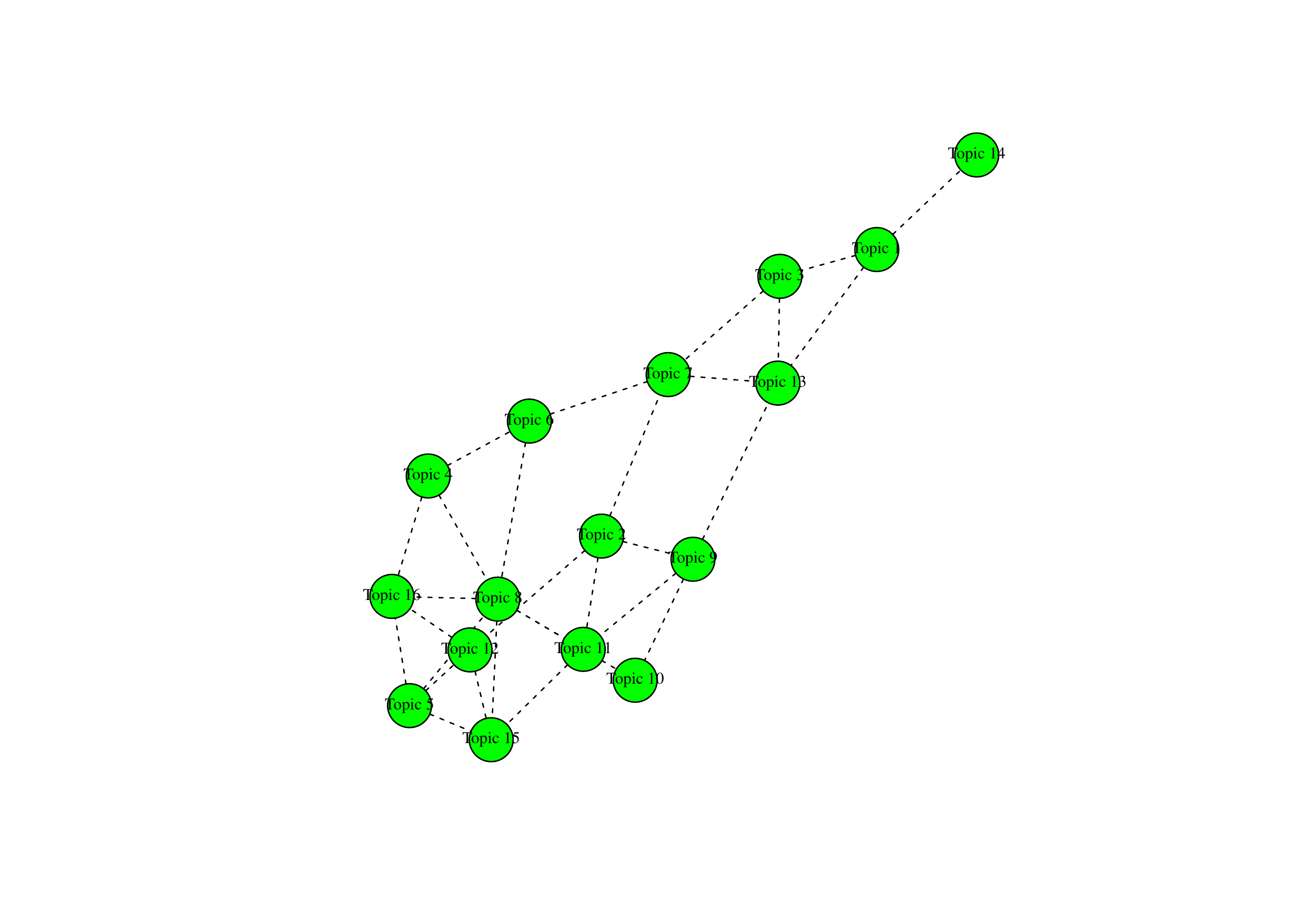} 
\caption{\emph{Network of topics}. Correlation of topics. 
\label{fig:correlation}}
\end{figure}

We also assess the relationship between topics in the STM framework, which allows correlations between topics to be examined. This is shown in the network of topics in Figure \ref{fig:correlation}. The figure shows that Topic 2 and Topic 7 are closely related, which we would expect as they both deal with international development (and share key words on development, such as `develop-', `povert-', etc.). It is also worth noting that while Topic 2 is more closely correlated with the Latin America topic (Topic 11), Topic 7 is more directly correlated with the Africa topic (Topic 6). 

%%%%%%%%%%%%%%%%%%%%%%%%%%%%%%%%%%%%%%%%
%%%%%%%%%%%%%%%%%%%%%%%%%%%%%%%%%%%%%%%%

\section{Explaining the rhetoric}

We next look at the relationship between topic proportions and structural factors. The data for these structural covariates is taken from the World Bank's World Development Indicators (WDI) unless otherwise stated. Confidence intervals produced by the method of composition in STM allow us to pick up statistical uncertainty in the linear regression model. 

\begin{figure}
\centering

\includegraphics[width=.45\textwidth]{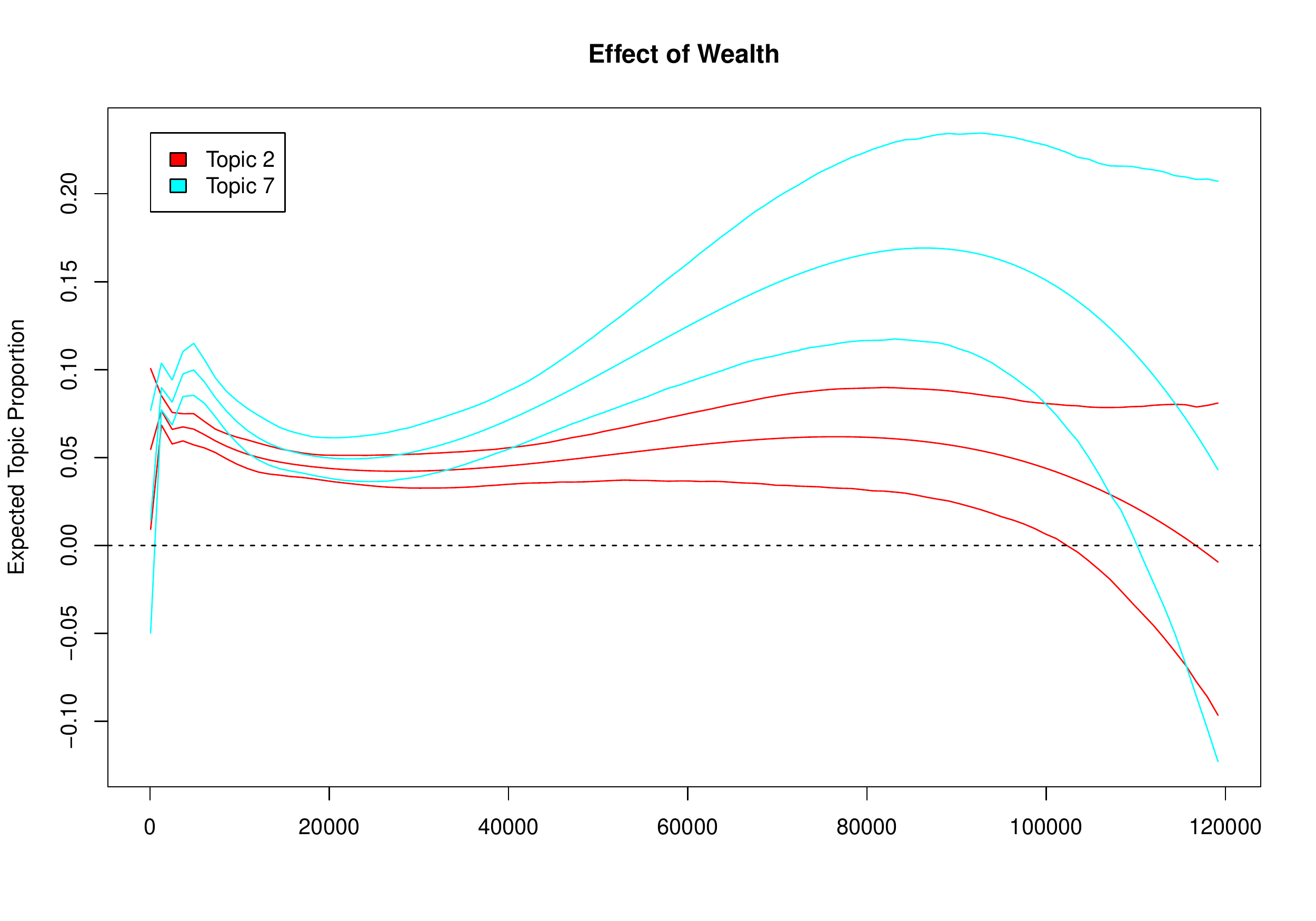} 
\caption{\emph{Effect of wealth}. Main effect and 95\% confidence interval. 
\label{fig:wealth}}
\end{figure}

Figure \ref{fig:wealth} demonstrates the effect of wealth (GDP per capita) on the the extent to which states discuss the two international development topics in their GD statements. The figure shows that the relationship between wealth and the topic proportions linked to international development differs across Topic 2 and Topic 7. Discussion of Topic 2 (economic development) remains far more constant across different levels of wealth than Topic 7. The poorest states tend to discuss both topics more than other developing nations. However, this effect is larger for Topic 7. There is a decline in the proportion of both topics as countries become wealthier until around \$30,000 when there is an increase in discussion of Topic 7. There is a further pronounced increase in the extent countries discuss Topic 7 at around \$60,000 per capita. However, there is a decline in expected topic proportions for both Topic 2 and Topic 7 for the very wealthiest countries.

%%%%%%%%%%%%%%%%%%%%%%%%%%%%%%%%%%%%%%%%
%%%%%%%%%%%%%%%%%%%%%%%%%%%%%%%%%%%%%%%%

\begin{figure}
\centering

\includegraphics[width=.45\textwidth]{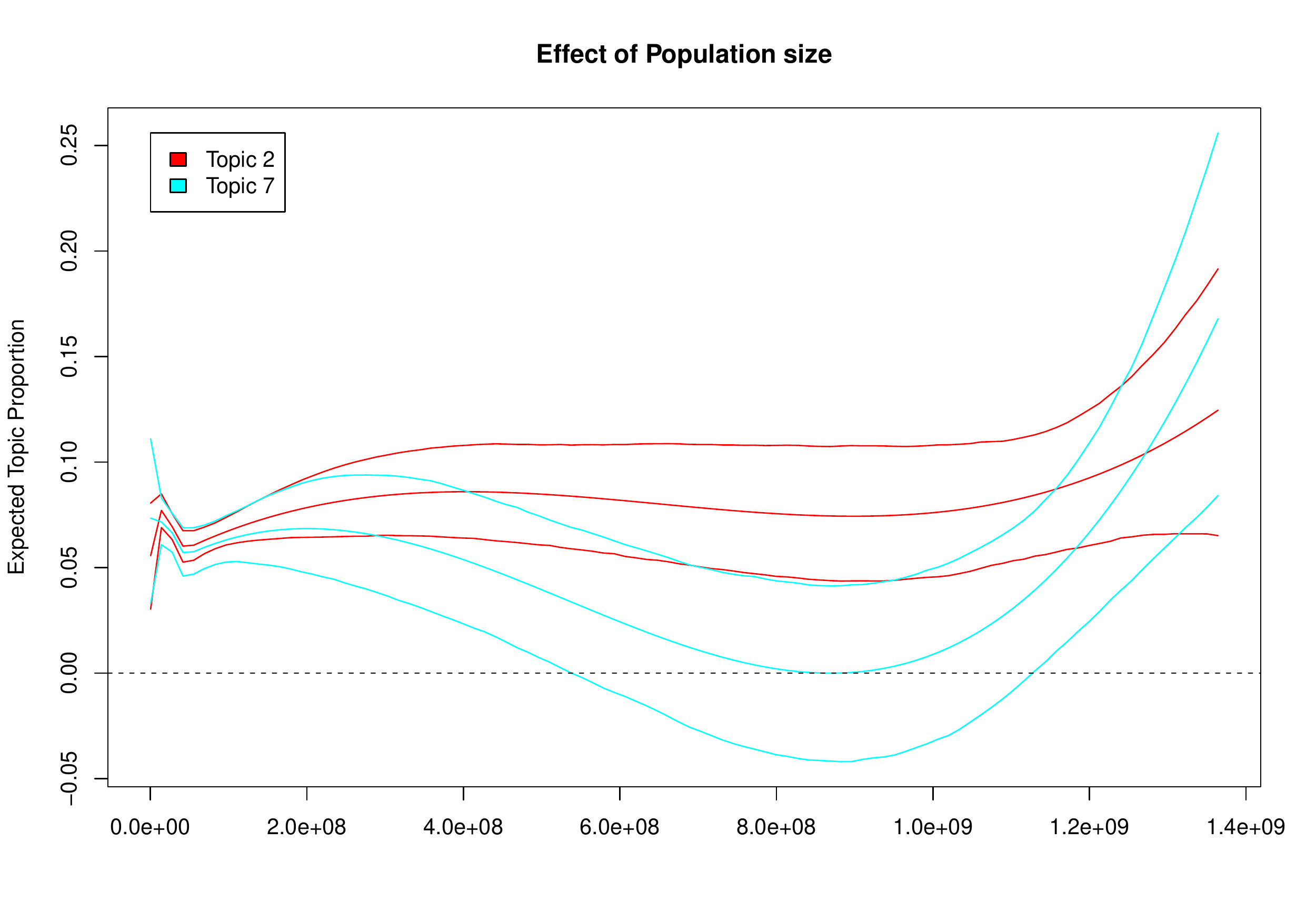} 
\caption{\emph{Effect of population}. Main effect and 95\% confidence interval. 
\label{fig:population}}
\end{figure}

Figure \ref{fig:population} shows the expected topic proportions for Topic 2 and Topic 7 associated with different population sizes. The figure shows a slight surge in the discussion of both development topics for countries with the very smallest populations. This reflects the significant amount of discussion of development issues, particularly sustainable development (Topic 7) by the small island developing states (SIDs). The discussion of Topic 2 remains relatively constant across different population sizes, with a slight increase in the expected topic proportion for the countries with the very largest populations. However, with Topic 7 there is an increase in expected topic proportion until countries have a population of around 300 million, after which there is a decline in discussion of Topic 7. For countries with populations larger than 500 million there is no effect of population on discussion of Topic 7. It is only with the very largest populations that we see a positive effect on discussion of Topic 7.

%%%%%%%%%%%%%%%%%%%%%%%%%%%%%%%%%%%%%%%%
%%%%%%%%%%%%%%%%%%%%%%%%%%%%%%%%%%%%%%%%

\begin{figure}
\centering

\includegraphics[width=.45\textwidth]{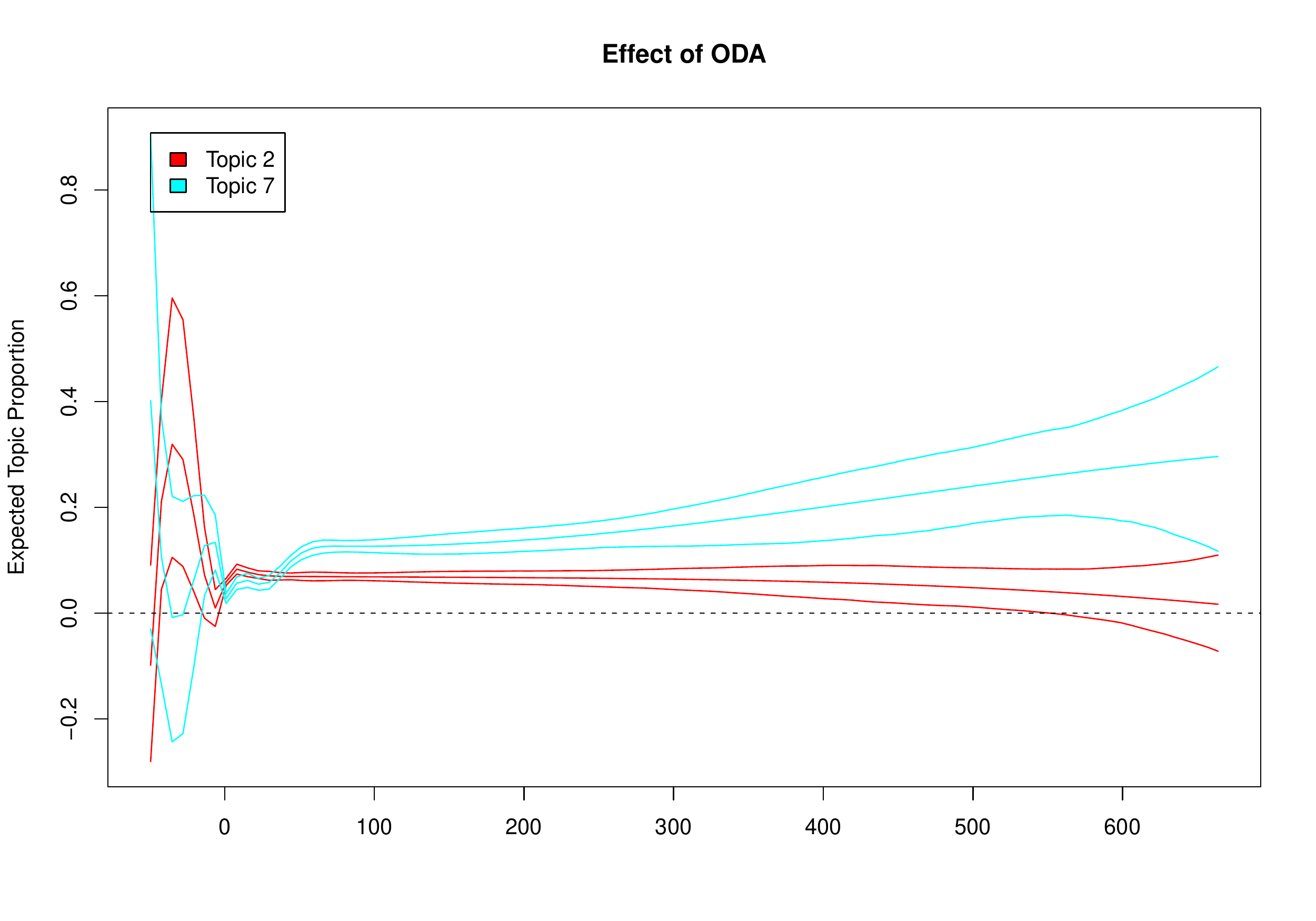} 
\caption{\emph{Effect of ODA}. Main effect and 95\% confidence interval. 
\label{fig:oda}}
\end{figure}

We would also expect the extent to which states discuss international development in their GD statements to be impacted by the amount of aid or official development assistance (ODA) they receive. Figure \ref{fig:oda} plots the expected topic proportion according to the amount of ODA countries receive. Broadly-speaking the discussion of development topics remains largely constant across different levels of ODA received. There is, however, a slight increase in the expected topic proportions of Topic 7 according to the amount of ODA received. It is also worth noting the spikes in discussion of Topic 2 and Topic 7 for countries that receive negative levels of ODA. These are countries that are effectively repaying more in loans to lenders than they are receiving in ODA. These countries appear to raise development issues far more in their GD statements, which is perhaps not altogether surprising.

%%%%%%%%%%%%%%%%%%%%%%%%%%%%%%%%%%%%%%%%
%%%%%%%%%%%%%%%%%%%%%%%%%%%%%%%%%%%%%%%%

\begin{figure}
\centering

\includegraphics[width=.45\textwidth]{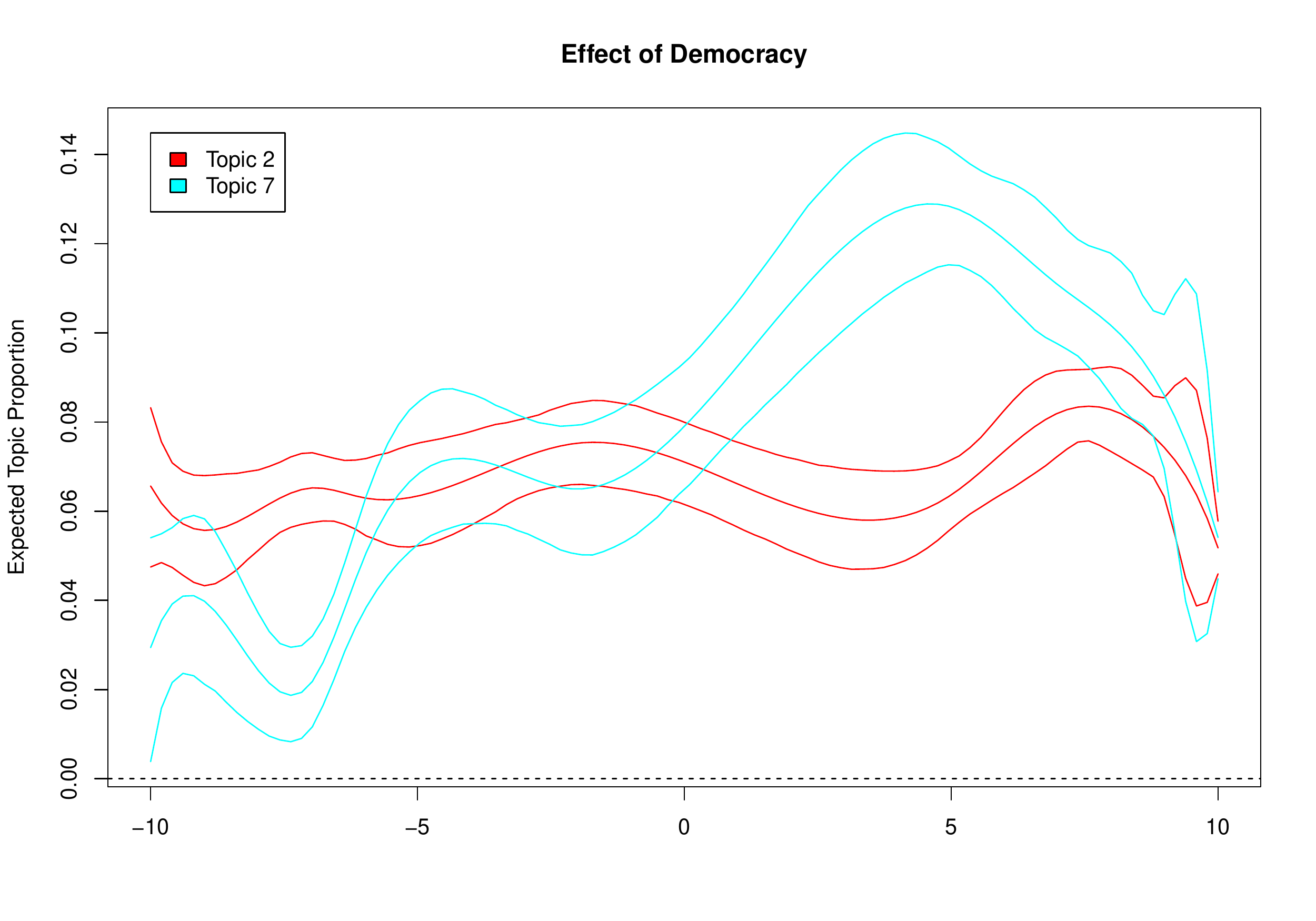} 
\caption{\emph{Effect of democracy}. Main effect and 95\% confidence interval. 
\label{fig:polity}}
\end{figure}

We also consider the effects of democracy on the expected topic proportions of both development topics using the Polity IV measure of democracy \cite{polity2003}. Figure \ref{fig:polity} shows the extent to which states discuss the international development topics according to their level of democracy. Discussion of Topic 2 is fairly constant across different levels of democracy (although there are some slight fluctuations). However, the extent to which states discuss Topic 7 (sustainable development) varies considerably across different levels of democracy. Somewhat surprisingly the most autocratic states tend to discuss Topic 7 more than the slightly less autocratic states. This may be because highly autocratic governments choose to discuss development and environmental issues to avoid a focus on democracy and human rights. There is then an increase in the expected topic proportion for Topic 7 as levels of democracy increase reaching a peak at around 5 on the Polity scale, after this there is a gradual decline in discussion of Topic 7. This would suggest that democratizing or semi-democratic countries (which are more likely to be developing countries with democratic institutions) discuss sustainable development more than established democracies (that are more likely to be developed countries). 

%%%%%%%%%%%%%%%%%%%%%%%%%%%%%%%%%%%%%%%%
%%%%%%%%%%%%%%%%%%%%%%%%%%%%%%%%%%%%%%%%

\begin{figure}
\centering

\includegraphics[width=.45\textwidth]{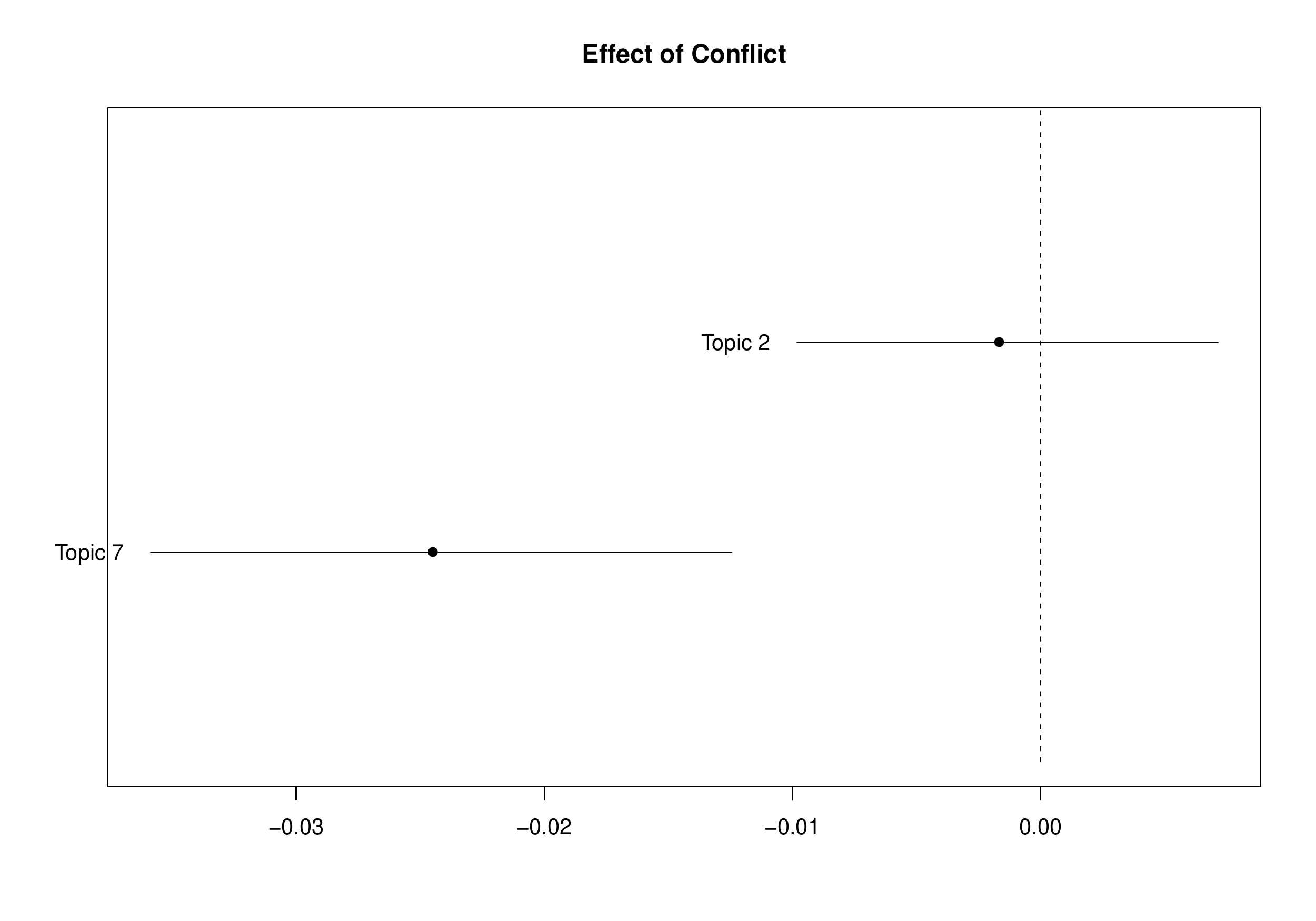} 
\caption{\emph{Effect of conflict}. Point estimates and 95\% confidence intervals.
\label{fig:conflict}}
\end{figure}

We also plot the results of the analysis as the difference in topic proportions for two different values of the effect of conflict. Our measure of whether a country is experiencing a civil conflict comes from the UCDP/PRIO Armed Conflict Dataset \cite{gleditsch2002armed}.  Point estimates and 95\% confidence intervals are plotted in Figure \ref{fig:conflict}.  The figure shows that conflict affects only Topic 7 and not Topic 2. Countries experiencing conflict are less likely to discuss Topic 7 (sustainable development) than countries not experiencing conflict. The most likely explanation is that these countries are more likely to devote a greater proportion of their annual statements to discussing issues around conflict and security than development. The fact that there is no effect of conflict on Topic 2 is interesting in this regard.

%%%%%%%%%%%%%%%%%%%%%%%%%%%%%%%%%%%%%%%%
%%%%%%%%%%%%%%%%%%%%%%%%%%%%%%%%%%%%%%%%

\begin{figure}
\centering

\includegraphics[width=.45\textwidth]{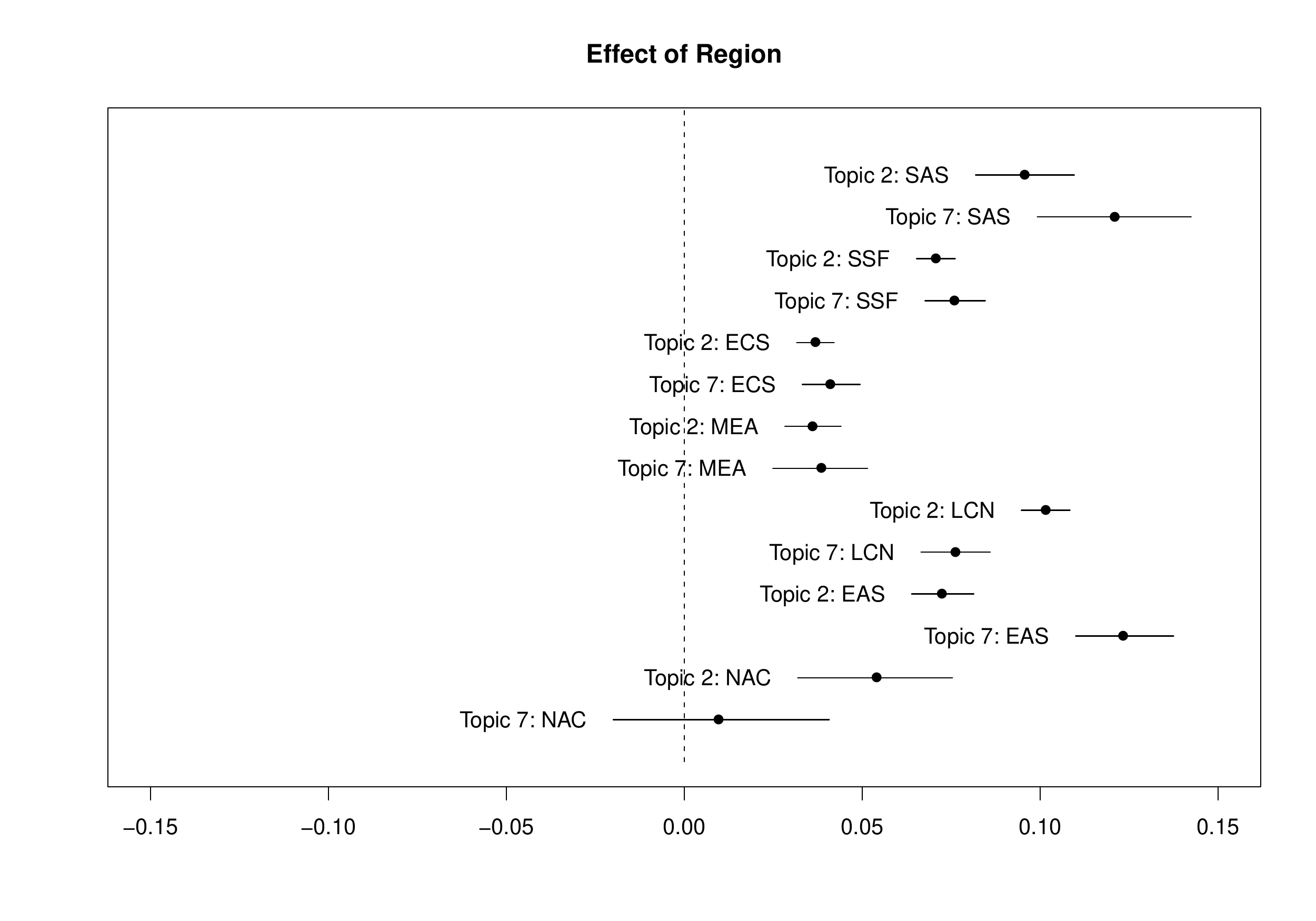} 
\caption{\emph{Regional effects}. Point estimates and 95\% confidence intervals. 
\label{fig:region}}
\end{figure}

Finally, we consider regional effects in Figure \ref{fig:region}. We use the World Bank's classifications of regions: Latin America and the Caribbean (LCN), South Asia (SAS), Sub-Saharan Africa (SSA), Europe and Central Asia (ECS), Middle East and North Africa (MEA), East Asia and the Pacific (EAS), North America (NAC). The figure shows that states in South Asia, and Latin America and the Caribbean are likely to discuss Topic 2 the most. States in South Asia and East Asia and the Pacific discuss Topic 7 the most. The figure shows that countries in North America are likely to speak about Topic 7 least.

%%%%%%%%%%%%%%%%%%%%%%%%%%%%%%%%%%%%%%%%
%%%%%%%%%%%%%%%%%%%%%%%%%%%%%%%%%%%%%%%%

The analysis of discussion of international development in annual UN General Debate statements therefore uncovers two principle development topics: economic development and sustainable development. We find that discussion of Topic 2 is not significantly impacted by country-specific factors, such as wealth, population, democracy, levels of ODA, and conflict (although there are regional effects). However, we find that the extent to which countries discuss sustainable development (Topic 7) in their annual GD statements varies considerably according to these different structural factors. The results suggest that broadly-speaking we do not observe linear trends in the relationship between these country-specific factors and discussion of Topic 7. Instead, we find that there are significant fluctuations in the relationship between factors such as wealth, democracy, etc., and the extent to which these states discuss sustainable development in their GD statements. These relationships require further analysis and exploration.

\section{Conclusion}

Despite decisions taken in international organisations having a huge impact on development initiatives and outcomes, we know relatively little about the agenda-setting process around the global governance of development. Using a novel approach that applies NLP methods to a new dataset of speeches in the UN General Debate, this paper has uncovered the main development topics discussed by governments in the UN, and the structural factors that influence the degree to which governments discuss international development. In doing so, the paper has shed some light on state preferences regarding the international development agenda in the UN. The paper more broadly demonstrates how text analytic approaches can help us to better understand different aspects of global governance.

%\section*{Acknowledgment}
%
%
%The authors would like to thank...
%

% trigger a \newpage just before the given reference
% number - used to balance the columns on the last page
% adjust value as needed - may need to be readjusted if
% the document is modified later
%\IEEEtriggeratref{8}
% The "triggered" command can be changed if desired:
%\IEEEtriggercmd{\enlargethispage{-5in}}

% references section

% can use a bibliography generated by BibTeX as a .bbl file
% BibTeX documentation can be easily obtained at:
% http://mirror.ctan.org/biblio/bibtex/contrib/doc/
% The IEEEtran BibTeX style support page is at:
% http://www.michaelshell.org/tex/ieeetran/bibtex/
%\bibliographystyle{IEEEtran}
% argument is your BibTeX string definitions and bibliography database(s)
%\bibliography{undevelopment}

\bibliographystyle{IEEEtran}
\bibliography{IEEEabrv,undevelopment}

%
% <OR> manually copy in the resultant .bbl file
% set second argument of \begin to the number of references
% (used to reserve space for the reference number labels box)
%\begin{thebibliography}{1}
%
%\bibitem{IEEEhowto:kopka}
%H.~Kopka and P.~W. Daly, \emph{A Guide to \LaTeX}, 3rd~ed.\hskip 1em plus
%  0.5em minus 0.4em\relax Harlow, England: Addison-Wesley, 1999.
%
%\end{thebibliography}

% that's all folks
\end{document}